\documentclass{article}

\usepackage[preprint]{neurips_2025}

\newif\ifarxiv
\arxivtrue

\usepackage[utf8]{inputenc} 
\usepackage[T1]{fontenc}    
\usepackage{hyperref}       
\usepackage{url}            
\usepackage{booktabs}       
\usepackage{amsfonts}       
\usepackage{nicefrac}       
\usepackage{microtype}      
\usepackage{xcolor}         
\usepackage{graphicx}
\usepackage{amsmath}
\usepackage{algorithm}
\usepackage{algpseudocode}
\usepackage{multicol}
\usepackage{subcaption}
\usepackage{wrapfig}
\usepackage{arydshln}  
\usepackage{titlesec}
\usepackage{caption}

\usepackage{setspace}
\titlespacing*{\section}
  {0pt}{2.5ex plus .2ex minus .2ex}{1.5ex plus .2ex}
\titlespacing*{\subsection}
  {0pt}{2.0ex plus .2ex minus .2ex}{1.0ex plus .2ex}

\titlespacing*{\section}
  {0pt}
  {1ex plus .2ex minus .2ex}
  {1ex plus .2ex}

\title{Latent Flow Transformer}

\author{%
  \begin{tabular}[t]{c}
    Yen‐Chen Wu\thanks{These authors contributed equally.}%
    \quad Feng‐Ting Liao\footnotemark[1]%
    \quad Meng‐Hsi Chen \\[1ex]
    Pei‐Chen Ho \quad Farhang Nabiei \quad Da‐shan Shiu
  \end{tabular}
  \\[2ex] \\
  MediaTek Research\thanks{%
    Correspondence: \texttt{yen-chen.wu@mtkresearch.com}, \texttt{ft.liao@mtkresearch.com}%
  }
}

\begin{document}

\maketitle

\begin{abstract}

Transformers, the standard implementation for large language models (LLMs), typically consist of tens to hundreds of discrete layers. 
While more layers can lead to better performance, this approach has been challenged as far from efficient, especially given the superiority of continuous layers demonstrated by diffusion and flow-based models for image generation.
We propose the Latent Flow Transformer (LFT), which replaces a block of layers with a single learned transport operator trained via flow matching, offering significant compression while maintaining compatibility with the original architecture.
Additionally, we address the limitations of existing flow-based methods in \textit{preserving coupling} by introducing the Flow Walking (FW) algorithm. 
On the Pythia-410M model, LFT trained with flow matching compresses 6 of 24 layers and outperforms directly skipping 2 layers (KL Divergence of LM logits at 0.407 vs. 0.529), demonstrating the feasibility of this design. When trained with FW, LFT further distills 12 layers into one while reducing the KL to 0.736 surpassing that from skipping 3 layers (0.932), significantly narrowing the gap between autoregressive and flow-based generation paradigms.
\end{abstract}

\vspace{-0.2cm}
\section{Introduction}
\vspace{-0.2cm}


Transformer-based large language models (LLMs) have become foundational to progress in NLP. While their scale enables remarkable capabilities, it also introduces substantial memory and compute demands, limiting their accessibility. Although evidence suggests many layers are overparameterized \cite{zafrir2021prune}, mainstream compression techniques like pruning and knowledge distillation \cite{lu_reassessing_2024,xu2024surveyknowledgedistillationlarge} have yet to deliver the scale of reduction required to fully address these limitations.


In image generation, diffusion and flow matching (FM) models \cite{lipman2022flow} have demonstrated remarkable efficiency, with recent approaches generating high-quality images in as few as one sampling pass \cite{frans_one_2024}. Inspired by this success, several works have explored adapting FM and diffusion models for language modeling \cite{sun2019patientknowledgedistillationbert,xu2024surveyknowledgedistillationlarge}. While these approaches offer promising directions, their performance still trails behind SOTA conventional transformer models—likely due to the discrete and compositional nature of language, where token-level generation resists continuous approximation.


Recent studies frame transformer forward passes as discretized continuous dynamical processes \cite{chen2018neural,li-etal-2022-ode,feinashley2025flowinglayerscontinuousdynamical}, and show that latent trajectories across layers can be nonlinear and complex. Concurrently, in image modeling, it has been shown that efficient surrogate models can be trained to approximate the latent mapping of deep networks using straighter transport dynamics \cite{das2023imagegenerationshortestpath,song2023consistencymodels}. These developments motivate exploring whether flow-based approaches can learn more efficient inter-layer mappings in transformers, offering a new path to structural compression.


Building upon these insights, We propose the \textbf{Latent Flow Transformer (LFT)}, a  novel transformer variant designed for parameter-and-compute efficient language modeling by replacing a contiguous block of transformer layers with a single learned transport operator trained via flow matching \cite{lipman2022flow}. To guide which layers are suitable for replacement, we introduce the \textit{Recoupling Ratio}, an interpretable Optimal Transport-based metric that accurately predicts flow matching quality. Our analysis finds that middle layers are particularly compressible, supporting the intuition that early LLM layers function qualitatively differently from later ones.


In our experiments training the LFT, with only $10^8$ training tokens, we show that standard flow matching enables effective LFT training, with perturbations smaller than those caused by removing a single transformer layer \cite{Sajjad_2023}. However, performance plateaus as flow path proximity introduces velocity ambiguity \cite{kim_simulation-free_2024,park2024constantaccelerationflow}—a known limitation in FM models. While reflow techniques are a common strategy to mitigate this issue, they are inherently incompatible with the LFT's objective of preserving the original input-output latent relationship.


To address this limitation, while advanced flow matching techniques offer potential solutions, we introduce \textbf{Flow Walking (FW)}, a scalable algorithm for learning latent transport using numerical integration. By training a transformer-like policy to explore paths with clearer view to the target, FW enables the model to navigate and resolve local ambiguities arising from multiple nearby flow paths, thereby significantly improving transport \textit{alignment} and overcoming the performance plateau observed with standard flow matching. 

We conduct extensive experiments on the \textit{Pythia-410M} model to validate the proposed framework. First, we find that layer selection is critical: LFTs constructed by replacing the layers identified by the \textit{Recoupling Ratio} consistently outperform those using arbitrary layer selections. Second, we demonstrate that LFT trained via flow matching can compress 12 out of 24 layers into a single transport layer while achieving a lower performance drop (KL divergence of LM logits at 0.407) than simply skipping two layers (KL divergence of LM logits at 0.529), thereby establishing the feasibility of our architecture. Finally, when trained with FW under the same compression setting, the LFT achieves a substantially lower KL divergence of LM logits at 0.736 than that with skipping three layers at 0.932, overcoming the primary obstacle in aligning latent transport across distant transformer layers and effectively bridging autoregressive and flow-based modeling paradigms.

In summary, our contributions are as follows.
\begin{itemize}
\item We introduce the Latent Flow Transformer, a novel architecture for efficient language modeling that replaces a block of transformer layers with a single learned transformer-like layer based on flow-matching principles.
\item We develop an interpretable predictor \textit{Recoupling Ratio} that estimates the suitability of different transformer layers for replacement. Our analysis with the predictor suggests that the middle layers of transformer models are particularly amenable to this form of compression, supporting hypotheses about the differentiated roles of layers in LLMs.
\item We demonstrate that while standard flow matching can learn the LFT's transport operator with a modest data budget, it encounters a performance plateau stemming from challenges such as flow path proximity in the latent space.
\item We propose Flow Walking, a scalable explorative algorithm for learning latent transport trajectories, designed to effectively navigate and resolve flow path proximity issues.
\item We empirically validate the LFT trained with both FM and FW on the \textit{Pythia}-410M model, demonstrating substantial parameter reduction with limited initial performance degradation and significant capability recovery after fine-tuning.
\end{itemize}

We release our code for the community to reproduce our results at \url{https://github.com/mtkresearch/latent-flow-transformer}


\vspace{-0.2cm}
\section{Preliminary}
\vspace{-0.2cm}

\subsection{Continuous-Time perspective of transformer layers}
\vspace{-0.2cm}

A neural net with $L$ discrete layers, especially with a residual connection, can be seen as refining its input hidden state $h_{l-1}$ at layer $l$, $l = 1, 2, ..., L$, to a better state at its output, $h_{l}$. A continuous-time counterpart can be expressed by the following dynamics ~\cite{chen2018neural,chen2023contiformer,papamakarios2021normalizing}:
{
  \setlength{\abovedisplayskip}{3pt}
  \setlength{\belowdisplayskip}{3pt}
  \begin{equation}\label{eq:definition_of_velocity}
    \scalebox{0.9}{%
      $\displaystyle
        v_t = \frac{d h_t}{d t} = u_{\theta}(h_t, t)
      $
    }
  \end{equation}
}
%
%
where $\theta$ parameterizes the ordinary differential equation (ODE), and $h_t$ matches with $h_l$ whenever $t$ matches $l/L$. Learning $\theta$ involves conducting expensive simulation ~\cite{kidger2022neural}:
%
{
  \setlength{\abovedisplayskip}{3pt}
  \setlength{\belowdisplayskip}{3pt}
  \begin{equation}\label{eq:integral}
    \scalebox{0.9}{%
      $\displaystyle
        h_{t_2} = h_{t_1}
        + \int_{t_1}^{t_2} u_{\theta}(h_t, t)\,dt
      $
    }
  \end{equation}
}

\subsection{Flow Matching for Simulation-Free Training}
\label{sec:fm_for_sim_free_training}
\vspace{-0.2cm}

Flow-based approaches~\cite{albergo2022building,lipman2022flow,liu2022rectified} provide a simulation-free solution to the ODE defined in \autoref{eq:definition_of_velocity}.
Flow matching moves a particle at initial position $x_0$ to its target position $x_1$ via a predefined path in unit time, and learns the corresponding velocity field ~\cite{lipman2022flow} conditioned on a set of $(x_0, x_1)$ pairs by minimizing the following loss,
{
  \setlength{\abovedisplayskip}{3pt}
  \setlength{\belowdisplayskip}{3pt}
    \begin{equation} \label{eq:fm_loss}
        \mathcal{L}_{FlowMatching} = \mathbb{E}_t \left[ \| u_{\theta}(x_t, t) - v_t \|^2 \right],
    \end{equation}
}
where $x_t$ and $v_t$ are the position and the velocity of the particle at time $t$, respectively. 

A straight line, constant speed trajectory is a popular and practical choice for flow matching. With such a target trajectory, $x_t$ and $v_t$ are simply 
$x_t = (1-t) \cdot x_0 + t \cdot x_1$, and $v_t = x_1 -x_0$.

During discrete-time inference, datapoints are transported along the flow in discrete steps. 
When moving from time points $t$ to $t+d$, a step is:
{
  \setlength{\abovedisplayskip}{3pt}
  \setlength{\belowdisplayskip}{3pt}
\begin{equation}\label{eq:take_one_step}
x_{t+d} = x_t + d \cdot u_\theta(x_t, t).
\end{equation}
}
To enhance the stability and the accuracy of the flow trajectory, an alternative is to set the velocity at the midpoint of the currently estimated forward step:
{
  \setlength{\abovedisplayskip}{3pt}
  \setlength{\belowdisplayskip}{3pt}
\begin{equation} \label{eq:take_one_recursive_step}
x_{t+d} = x_t + d \cdot u_\theta(x_t + \frac{d}{2} \cdot u_\theta(x_t,t), t + \frac{d}{2}).
\end{equation}
}

\subsection{Challenges in Flow Matching for Paired Data}
\label{sec:paired-data-challenge}
\vspace{-0.2cm}

In flow matching with paired samples, it is crucial to preserve the deterministic correspondence between the support of the source and target distributions. Such setting, where interpolation trajectories intersect, challenges standard flow matching methods \cite{lipman2022flow,liu2022rectified} as these methods tend to average conflicting velocity signals near these intersections, yielding biased estimates and failing to transport source points accurately to their paired targets \cite{park2024constantaccelerationflow,kim_simulation-free_2024,frans_one_2024}. To overcome this, Work by \cite{park2024constantaccelerationflow} introduces an auxiliary acceleration field that regulates the rate of change along each trajectory, thereby sharpening the alignment to the true paired mapping. Alternatively,  \cite{kim2024simulation} propose learning latent-space projections that untangle intersecting pairs prior to flow estimation, effectively preventing trajectory crossings before velocity estimation.

\section{Latent Flow Transformer}
\label{sec:latent-flow-transformer}
\vspace{-0.2cm}

We propose the \textbf{Latent Flow Transformer (LFT)}, a novel architecture for language modeling that seeks to leverage the benefits of flow-related concepts, as demonstrated in image generation, for transformer-based large language models (LLMs). The LFT is designed to potentially achieve advantages such as improved parameter efficiency, simulation-free training, and latent trajectory straightening.

The LFT reduces model size by replacing a contiguous block of transformer layers from the teacher with a single learned transport operator, called the \textbf{latent flow layer}. This operator is trained using flow concepts \cite{lipman2022flow} to accurately map the input latent at the input of the block to its corresponding output over a desired trajectory. 

With a suitable architectural choice, the learned operator can be unrolled over predefined intermediate time steps to effectively recover a transformer-like structure. Furthermore, the number of layers used for a given token can be determined at run time on a per-token basis, giving further flexibility for context dependent performance-computation trade-off.

\begin{figure}[h]
\vspace{-0.5cm}
    \centering
    \includegraphics[width=0.75\textwidth]{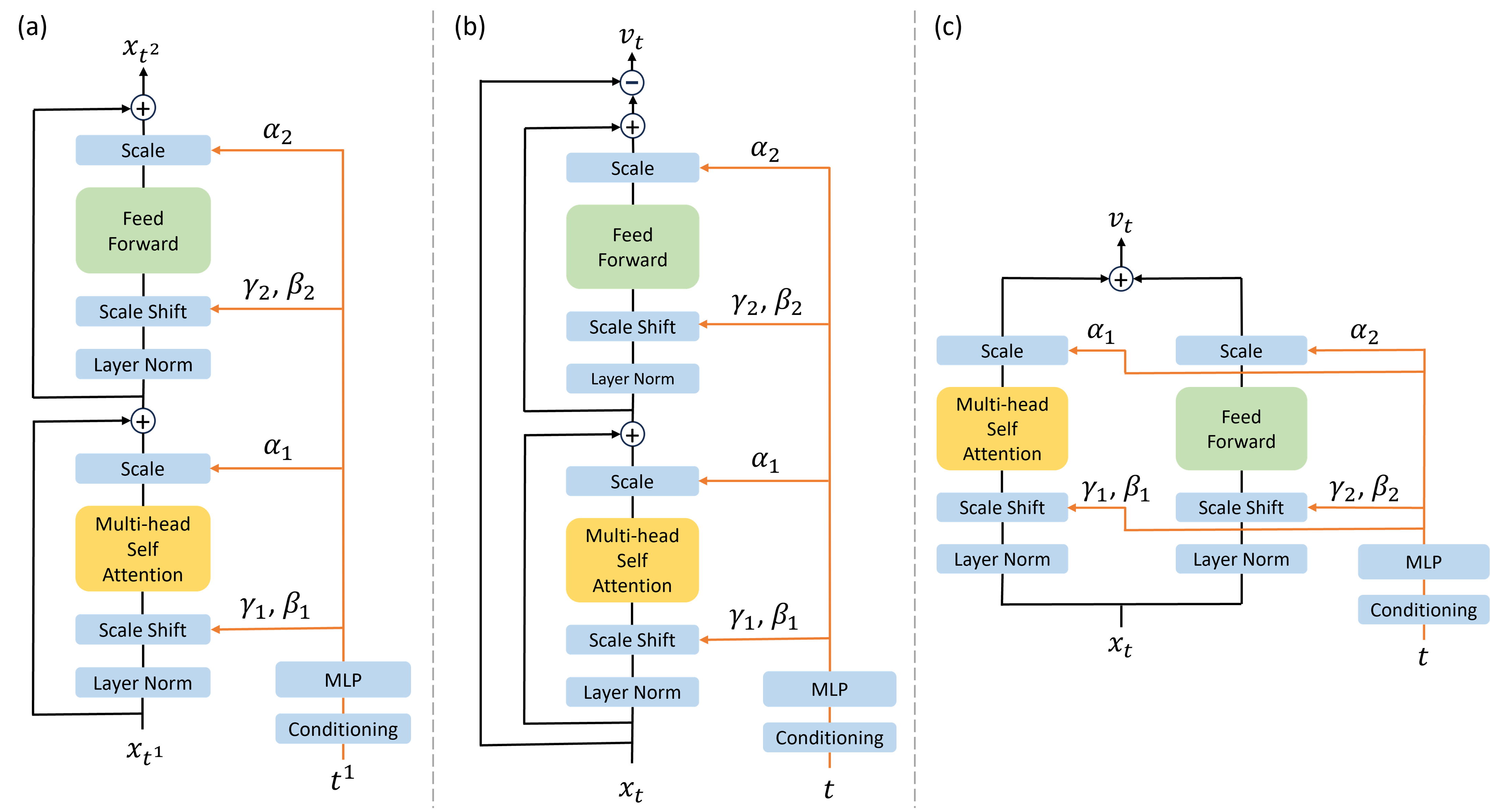}
    \caption{\textbf{Velocity field estimator.} (a) The DiT block, mapping an input hidden state to an output hidden state conditioned on time $t$. (b) Velocity field estimator derived from the DiT block, mapping an input state $x_t$ and time $t$ to a velocity $v_t$. (c) Velocity field estimator for \textit{Pythia}.} 
    \label{fig:dit-based-velocity-estimator}
 \vspace{-0.5cm}       
\end{figure}

\subsection{Velocity field estimating network}
\label{ssec:velocity-field-network}
\vspace{-0.2cm}

Specifying an LFT requires defining the network architecture for the velocity field estimator $u_\theta(x_t,t)$; following the approach in \cite{peebles2023scalablediffusionmodelstransformers}, our method augments a teacher transformer layer with additional scale and shift operators and an MLP predicting their factors, as shown in \autoref{fig:dit-based-velocity-estimator}(a). We derive the velocity estimate by subtracting the input latent from this augmented network's output (\autoref{fig:dit-based-velocity-estimator}(b)). \autoref{fig:dit-based-velocity-estimator}(c) details the specific case for \textit{Pythia} models. We remark that these networks are used solely for learning the velocity field; during LFT inference, they are augmented with skip connections to produce latents.

There are a lot of considerations and consequently room for innovation regarding how these estimators causally attend to prior tokens' latents. It is particularly interesting when evaluating $u_\theta(x_t,t)$ at different values of $t$ from token to token. We leave the details to the experiment section.

\subsection{Predict flow matching quality by the Recoupling ratio}
\label{ssec:recoupling-rate}
\vspace{-0.2cm}

Selecting the optimal block of layers for replacement is critical for LFT performance. Learnability via flow matching is bottlenecked by flow path crossing \cite{peyre2019computational,lipman2022flow}, a challenge exacerbated in LFT as preserving original input-output pairings precludes the use of reflow methods \cite{liu2022rectified,das2023imagegenerationshortestpath,kim_simulation-free_2024}. To guide layer selection, we introduce an estimator which we refer to as the \textit{Recoupling ratio}. Given sample latent pairs from layers $m$ to $n$, this estimator quantifies the deviation between their original pairing and the pairing dictated by Optimal Transport (OT). Optimal Transport identifies the minimal cost mapping between two distributions. In \autoref{eq:define-optimal-transport}, the OT matrix $M$ represents the ideal pairing between latents at layer $m$ and layer $n$ that minimizes a transport cost, such as Euclidean distance: 
{
  \setlength{\abovedisplayskip}{3pt}
  \setlength{\belowdisplayskip}{3pt}
  \begin{equation}\label{eq:define-optimal-transport}
    \scalebox{0.9}{%
      $\displaystyle
        M = \arg\min_{\gamma}
            \sum_{i,j} \gamma_{i,j}\,d\bigl(h_m^{(i)},\,h_n^{(j)}\bigr)
      $
    }
  \end{equation}
}
We define the Recoupling Ratio $R$ as the percentage of pairing relationship that disagrees with $M$. If $M$ is a square matrix of order $O_M$:
{
  \setlength{\abovedisplayskip}{3pt}
  \setlength{\belowdisplayskip}{3pt}
  \begin{equation}\label{eq:recoupling-ratio}
    \scalebox{0.9}{%
      $\displaystyle
        R := 1 - \mathbf{E}\bigl[\tfrac{\text{Tr}(M)}{O_M}\bigr].
      $
    }
  \end{equation}
}
Because the Recoupling Ratio quantifies the misalignment with original pairings, therefore, controlling for $O_M$, lower $R$ indicates better alignment and thus predicting fewer flow-crossing issues and higher feasibility for learning the LFT. In our experiments, a practically small $O_M$ is sufficient for the prediction to aid decision making on the selection of layers.

\subsection{Learning the velocity field}
\label{ssec:flow-matching-distillation}
\vspace{-0.2cm}



The velocity estimators shall learn the velocity field from a collection of $x_0, x_1$ pairs. For an LFT to replacing all layers from layer $m$ to layer $n$ (inclusive), omitting the details of preserving the shared context, we take the input latent of layer $m$ for a given token as $x_0$, and the output latent of layer $n$ as the corresponding $x_1$.   

To learn the velocity field via standard flow matching \cite{lipman2022flow}, Algorithm \ref{algo:standard_flow_matching} is applied. We remark that LFT can be trained by many other compatible flow-related algorithms.



\begin{center}
\vspace{-0.5cm}
\scalebox{0.8}{
\begin{minipage}{\textwidth} 
\begin{algorithm}[H]
    
\caption{Learning the velocity field via standard flow matching~\cite{lipman2022flow}}
\begin{algorithmic}[1]
\State $\mathcal{D}$ is a set of training data, $\text{LLM}_{\text{teacher}}(d,m)$ is the latent of LLM at layer $m$ over data $d$, $AdamW$ is the AdamW optimizer.
\While{termination condition not met}
    \State $d \sim \mathcal{D}$
    \State $(x_0, x_1) \leftarrow (\text{LLM}_{\text{teacher}}(d,m), \text{LLM}_{\text{teacher}}(d,n))$ 
    \State $t \sim U$
    \State $x_t \leftarrow (1 - t) x_0 + t x_1$ 
    \State $\theta \leftarrow AdamW(\theta, \nabla_{\theta} \| u_{\theta}(x_t, t) - (x_1 - x_0) \|^2$)
\EndWhile
\end{algorithmic}
\label{algo:standard_flow_matching}
\end{algorithm}
\end{minipage}
}
\end{center}

\subsection{Unrolled LFT has a Transformer structure}
\label{ssec:unroll-back-to-transformer}
\vspace{-0.2cm}

\begin{figure}[t!]
\vspace{-0.5cm}
\centering
\includegraphics[width=0.55\textwidth]{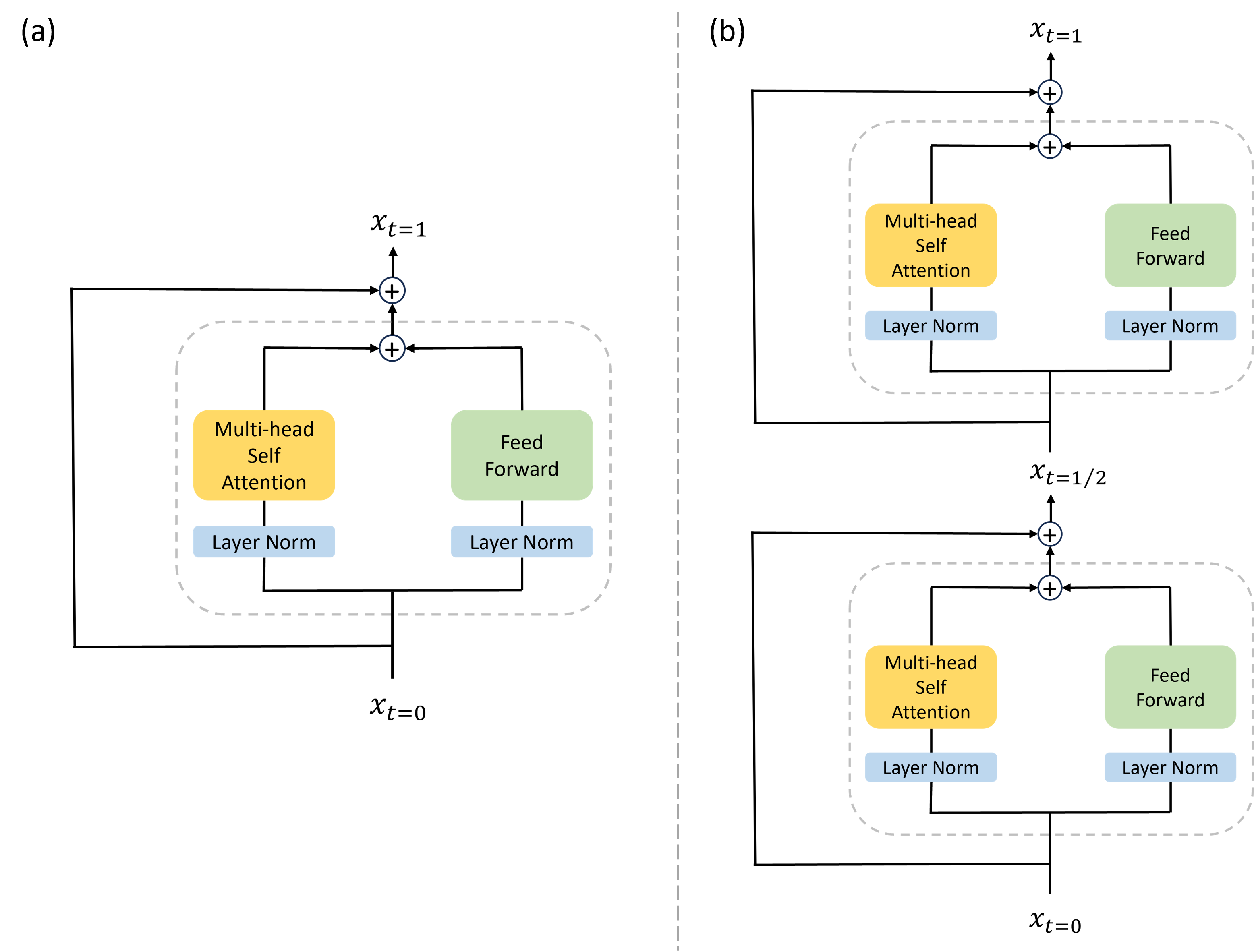}
\caption{\textbf{Static structure of an unrolled LFT.} We highlight only the latent flow layer. The simple reconstruction rule of \autoref{eq:take_one_step} is assumed. (a) LFT based on \textit{Pythia} with single step reconstruction. (b) LFT based on \textit{Pythia} with two step reconstruction. }
\label{fig:inference-single-and-two-steps}
\vspace{-0.5cm}
\end{figure}

At inference, unrolling the latent flow layer over a fixed set of time points $t_0 = 0 < t_1 < t_2 ... < 1$ hardens the latent evolution process into a static processing graph, beneficial for visualizing data flow and optimizing hardware implementation. For the specific case of a single step flow matching combined with the simple reconstruction approximation rule of \autoref{eq:take_one_step}, the latent flow layer becomes equivalent to a single standard transformer layer (see \autoref{fig:inference-single-and-two-steps}). If more than one step is used, the latent flow layer is equivalent to a stack of transformer layers with cross-layer attention~\cite{li2020cross}.

The structural similarity between the unrolled LFT and a standard transformer is of significant practical importance, enabling researchers and practitioners to leverage the extensive ecosystem and highly optimized infrastructure developed for transformer-based LLMs.

\vspace{-0.2cm}

\section{Flow Matching with Paired Data}
\label{sec:fw-paired-data-flow-matching}




\begin{figure}[h]
\vspace{-0.5cm}
    \centering
    \includegraphics[width=0.65\textwidth]{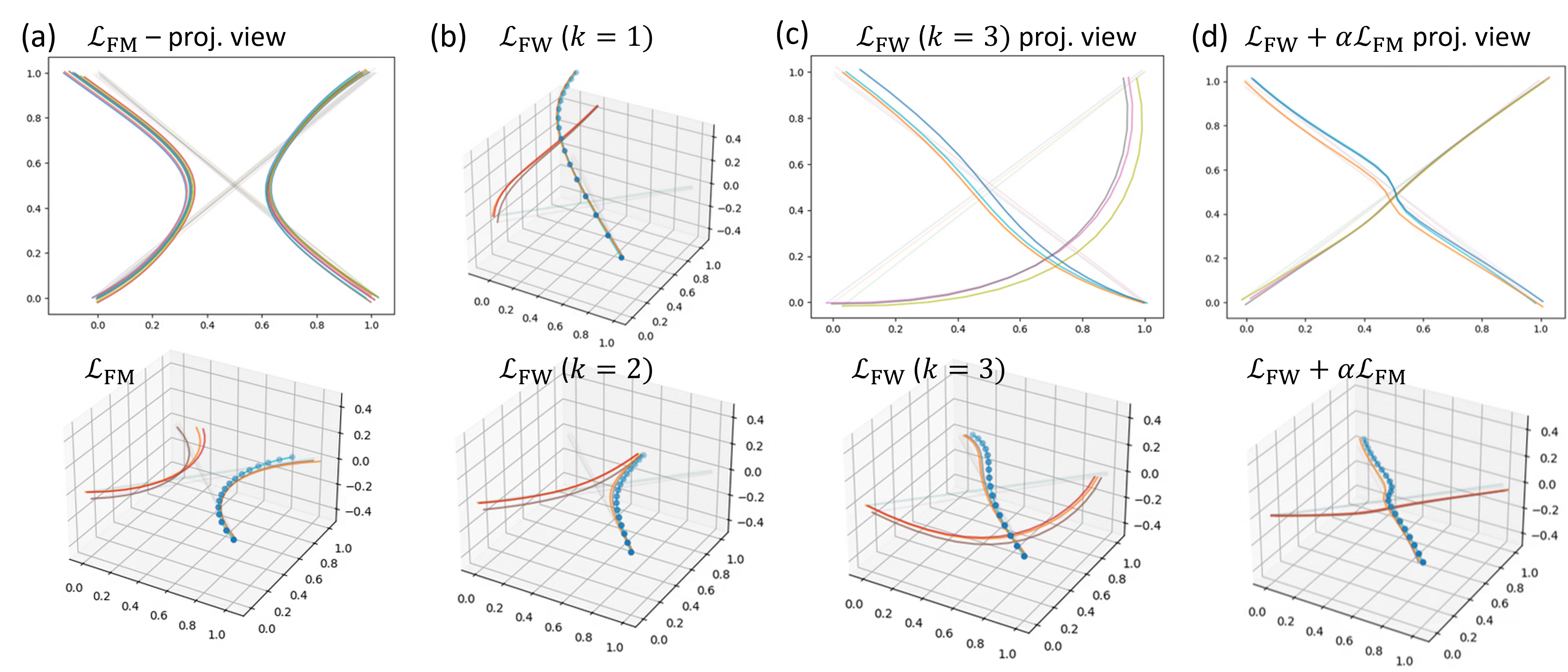}
    \caption{\textbf{Toy trajectories for paired data.} Faded lines show the ground‐truth trajectories of paired points, while solid curves depict predictions from various flow‐matching algorithms.  
    (a) Standard flow matching fails to maintain pairwise correspondence.  
    (b) FW results for $k=1$ (top) and $k=2$ (bottom): using fewer integration steps leads to poor trajectory estimates.  
    (c) FW with $k=3$: projected view (top) and full trajectories (bottom) demonstrate how the learned velocity field generates smooth, curved paths that avoid intersections.  
    (d) A hybrid of FW and standard flow matching with $\alpha=0.001$ preserves the constant velocity of paired data while allowing curvature at intersections to prevent flow crossings.}

    \label{fig:paired_fm}
\vspace{-0.5cm}    
\end{figure}

To address the challenge of crossing trajectory in \ref{sec:paired-data-challenge} while to enable the unrolling of latent flow layer back to a Transformer structure in \ref{ssec:unroll-back-to-transformer}, we introduce the Flow Walking (FW) algorithm for training and inference the latent flow layer of LFT.

FW approximates $x_0$ to $x_1$ using numerical integration with discrete time points. 
We define a step as $s_\theta(x_t,t,t') = x_{t+d}$, where $d=t'-t$ and  $x_{t+d}$ can be determined by \autoref{eq:take_one_step} or~\autoref{eq:take_one_recursive_step}.
Our key intuition is that we can learn non-crossing trajectories by slightly separate trajectories around intersecting points. We define the learning of velocity fields with:
\begin{equation}
  \setlength{\abovedisplayskip}{3pt}
  \setlength{\belowdisplayskip}{3pt}
  \scalebox{0.9}{%
    $\displaystyle
      \mathcal{L}_{\mathrm{FlowWalking}(k)}
      = \mathbb{E}_{t_{1},\dots,t_{k-1}}
      \biggl\|\,x_0 + \sum_{i=1}^k \Delta_{\theta, t_{i}} - x_1\biggr\|^2
    $,
  }
  \label{eq:FW_general_form}
\end{equation}
where $\Delta_{\theta, t_{i}} = s_\theta\bigl(x_{t_{i-1}},\,t_{i-1},t_{i}\bigr) - x_{t_{i-1}}$, $t_0=0$, $t_k=1$, and $t_i\sim[0,1]$ for $i\in[1,k-1]$. 
Unlike linear inference paths (\ref{sec:fm_for_sim_free_training}) that randomly samples different numbers of points $k$ \cite{geiping2025scaling}, we would like to choose the minimum $k$ with randomly sampled $t$ to avoid expensive simulation. 
We empirically choose an efficient and generalized training loss using $k=3$ (\autoref{fig:paired_fm}) as choosing $k=2$ and $t_1\sim[0,1]$ cannot generalize to inference time points $k>2$.


We compare FW with alternative training schemes on toy paired-data experiments (\autoref{fig:paired_fm}. In panel (a), standard flow matching - unstable for paired data - exhibits trajectory "recoupling,"  though in the absence of crossings it can recover a straight, constant speed path. Panels (b) and (c) show FW with $k=1$, and $k=2$ (top and bottom, respectively), which fail to learn a generalizable velocity field, and FW with $k=3$ preserves the original coupling at the expense of straightness. Finally, in panel (d), we add a straightness regularizer via standard flow matching, i.e.,
\begin{equation}
  \setlength{\abovedisplayskip}{3pt}
  \setlength{\belowdisplayskip}{3pt}
  \scalebox{0.9}{%
    $\displaystyle
    \mathcal{L}=\mathcal{L}_\text{FlowWalking}+\alpha\mathcal{L}_\text{FlowMatching}
    $,
  }
\end{equation}
which yields a non-crossing, straight interpolation of the paired data. While this stability-regularized variant shows promise for paired-data application, we leave its comprehensive study to future work and here focus on demonstrating FW’s feasibility to LFT.

\noindent
\makebox[\textwidth][c]{%
\scalebox{0.8}{%
\begin{minipage}[t]{0.58\textwidth}

\vspace{-0.5cm}
\begin{algorithm}[H]
\caption{LFT Training (FW Algorithm)}
\begin{algorithmic}[1]
\State $\mathcal{D}$ is a set of training data, $\text{LLM}_{\text{teacher}}(d,m)$ is the latent of LLM at layer $m$ over data $d$, $AdamW$ is the AdamW optimizer.
\While{termination condition not met}
    \State $d \sim \mathcal{D}$
    \State $(x_0, x_1) \leftarrow (\text{LLM}_{\text{teacher}}(d,m), \text{LLM}_{\text{teacher}}(d,n))$ 
    \State $t_1, t_2 \sim U$ \Comment{Sorted $t_2 > t_1$}
    \State $x_{t_1} \leftarrow s_{\theta}(x_0, 0, t_1)$ \Comment{1st step}
    \State $x_{t_2} \leftarrow s_{\theta}(x_{t_1}, t_1, t_2)$ \Comment{2nd step}
    \State $\hat{x}_1 \leftarrow s_{\theta}(x_{t_2}, t_2, 1)$ \Comment{3rd step}
    \State $\theta \leftarrow AdamW(\theta, \nabla_{\theta} \| \hat{x}_1 - x_1 \|^2$) 
\EndWhile
\end{algorithmic}
\label{algo:FW}
\end{algorithm}
\end{minipage}

\hspace{0.05\textwidth}\hfill\hspace{0.05\textwidth}
\begin{minipage}[t]{0.38\textwidth}
\vspace{-0.5cm}
\begin{algorithm}[H]
\caption{LFT Inference}
\begin{algorithmic}[1]
\State $x \leftarrow \text{EmbedIn}(token)$
\State $d \leftarrow 1 / k$
\State $t \leftarrow 0$
\State $t' \leftarrow d$
\For{$k \in [0, \ldots, k - 1]$}
    \State $x \leftarrow s_{\theta}(x, t, t')$
    \State $t \leftarrow t + d$
    \State $t' \leftarrow t' + d$
\EndFor
\State logits = $\text{EmbedOut}(x)$
\State \Return logits
\end{algorithmic}
\label{algo:LFT-inference}
\end{algorithm}
\end{minipage}%
}%
}

\section{Experiments}
\label{sec:exp}
\vspace{-0.2cm}

\begin{figure}[!t]
\centering
\begin{minipage}[t]{0.65\textwidth}
\vspace{0pt} 
\centering
\begin{subfigure}[b]{0.49\textwidth}
    \centering
    \includegraphics[width=\textwidth]{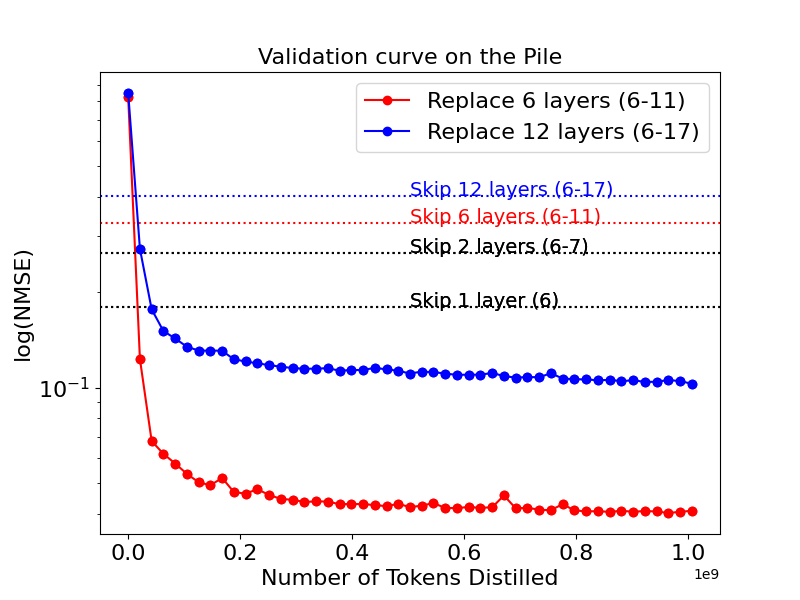}
    \caption{LFT-SFM}
\end{subfigure}
\hfill
\begin{subfigure}[b]{0.49\textwidth}
    \centering
    \includegraphics[width=\textwidth]{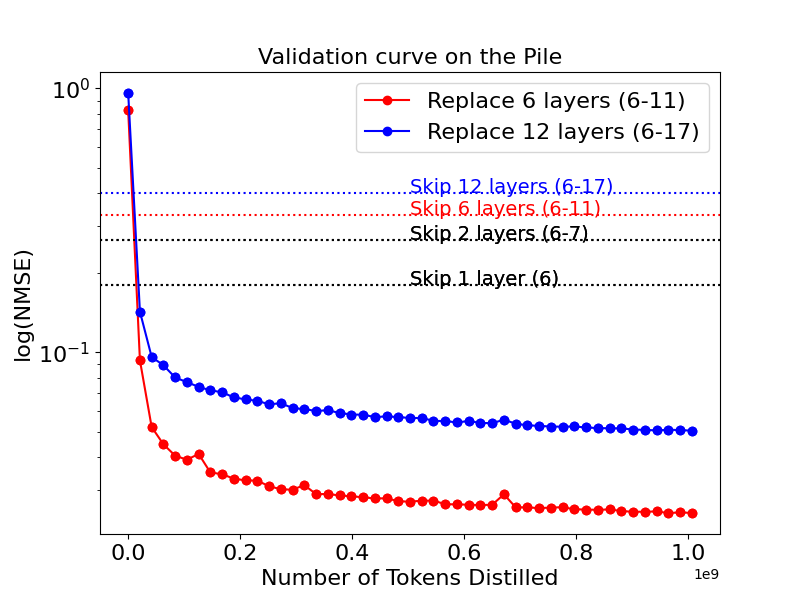}
    \caption{LFT-FW}
\end{subfigure}
\caption{\textbf{Latent distillation with flow matching vs layer-skipping baselines.} Validation NMSE between predicted and original hidden states vs. training tokens. The dotted lines show Pythia layer-skipping baselines. Flow-matching layers approximate original representations with lower reconstruction error than layer-skipping baselines across both setting.}
\label{fig:training_curve}
\end{minipage}
\hfill
\begin{minipage}[t]{0.31\textwidth}
\vspace{8pt} 
\centering
\includegraphics[width=\linewidth]{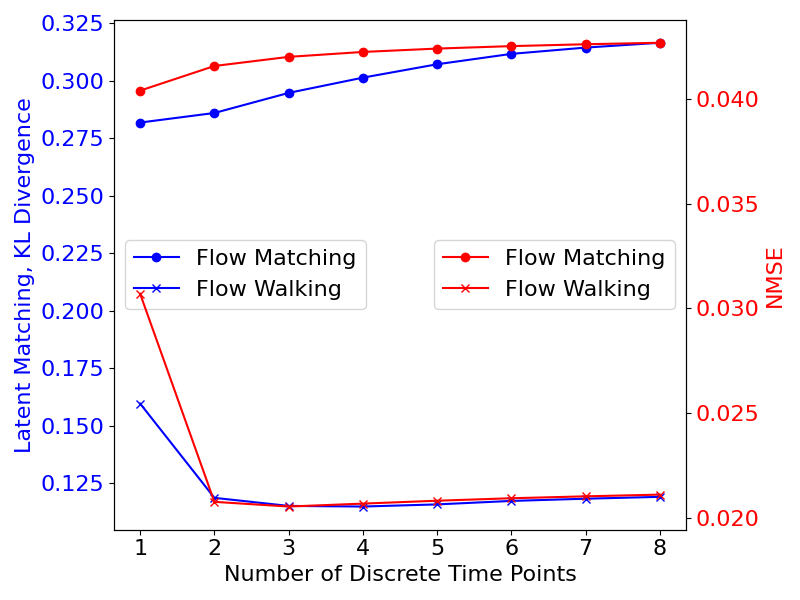} 
\caption{\textbf{Inference performance ($\textbf{KL}_{x||\hat{x}}$ and NMSE) vs number of discrete points $k$}. We report results of LFT-SFM and LFT-FW with latent flow layer replacing layer 6-18 of Pythia-410m.}
\label{fig:inference_metric_vs_step}
\end{minipage}
\vspace{-0.3cm}
\end{figure}

We implemented the LFT based on \texttt{Pythia-410m} and present a series of experiments evaluating the choice of distilled layers, the convergence of flow matching distillations, and inference as an integrated transformer. 
We implement Algorithm \ref{algo:standard_flow_matching} by following the reference code of \cite{lipman2024flowmatchingguidecode}, and to ensure a fair comparison between methods - we use the same recursive update rule \autoref{eq:take_one_recursive_step} to compute $x_{t+d}$ for FW.

\vspace{-0.2cm}
\subsection{Evaluation metric}
\vspace{-0.2cm}
To assess the local fidelity of a latent flow layer, we measure (i) the \textbf{normalized mean squared error} (NMSE) and (ii) the \textbf{KL divergence} between its output $\hat x_1$ and the target latent $x_1$, denoted as $\textbf{KL}_{x_1||\hat x_1}$.  The NMSE is defined as $\mathrm{NMSE} \;=\; \frac{\mathbb{E}\,\|\hat x_1 - x_1\|^2}{\mathbb{E}\,\|x_1\|^2}$.
%

For an end-to-end language-modeling evaluation, we compute the empirical KL divergence between the teacher distribution $P$ and the LFT output distribution $Q$, denoted as \textbf{$\textbf{KL}_{P||Q}$}, and we also report \textbf{perplexity} (PPL) based on the LFT’s output logits over a held-out subset of The Pile.

%

As a reference, Table \ref{table:comparison_selected} reports NMSE, KL divergence, and perplexity for the cases where we replace (a) layer 6, or (b) layers 6–7 of Pythia-410M with identity skip connections—conditions under which negligible degradation was previously observed \cite{gromov2025unreasonableineffectivenessdeeperlayers}.

\vspace{-0.2cm}
\subsection{Can Recoupling Ratio rank the feasibility of learning latent flow layers?}
\begin{figure}[htp]
\vspace{-0.5cm}
  \centering
  \begin{subfigure}[b]{0.28\textwidth}
    \centering
    \includegraphics[width=\textwidth]{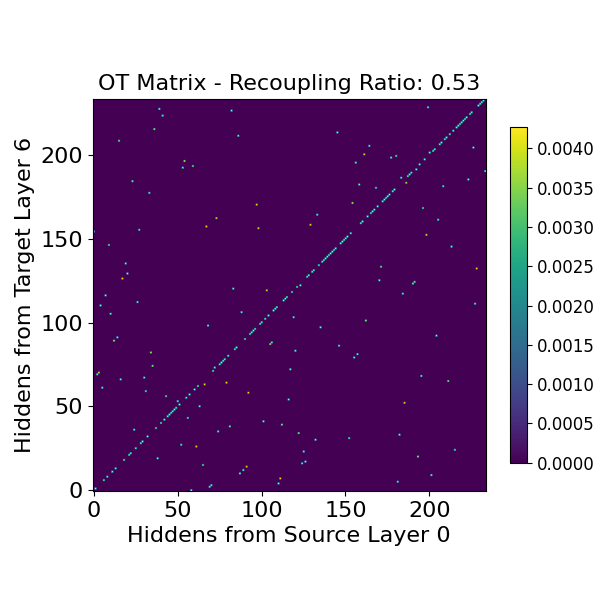}
    
    \caption{OT Matrix of paired hiddens between layer 0 and layer 6.}
    \label{fig:ot-matrix-0-6}
  \end{subfigure}
  \hfill
  \begin{subfigure}[b]{0.28\textwidth}
    \centering
    \includegraphics[width=\textwidth]{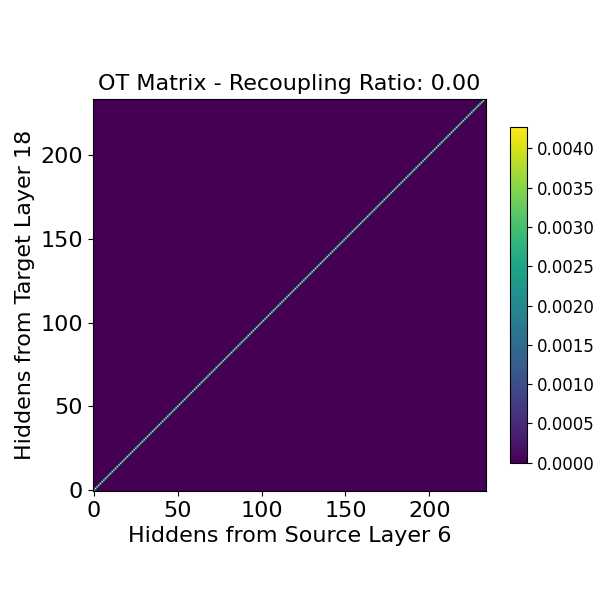}
    \caption{OT Matrix of paired hiddens between layer 6 and layer 18.}
    \label{fig:ot-matrix-6-18}
  \end{subfigure}
  \hfill
  \begin{subfigure}[b]{0.28\textwidth}
    \centering
    \includegraphics[width=\textwidth]{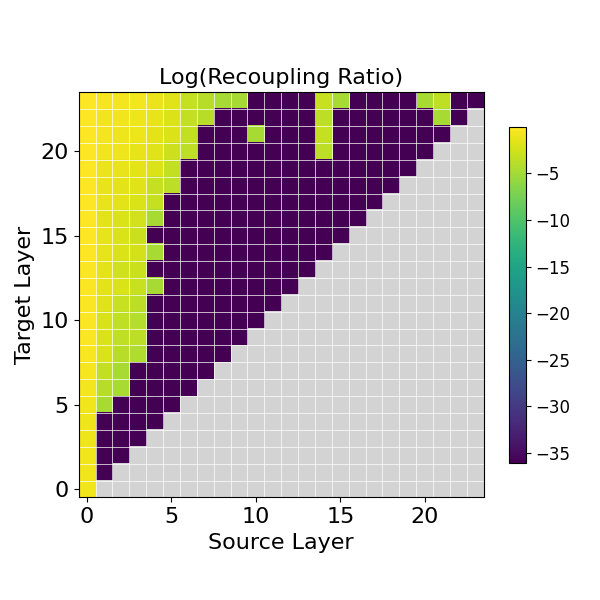}
    \caption{Recoupling Ratios of different source target layers.}
    \label{fig:recoupling-ratios-matrix}
  \end{subfigure}

  \caption[Optimal ransport and recoupling]{%
    \textbf{Optimal Transport Matrices and Recoupling Ratio.}
  }
    \label{fig:recoupling_rate}
    \vspace{-0.2cm}
\end{figure}

\vspace{-0.2cm}
Sampling just 256 tokens from the first portion of the Pile, we computed the recoupling matrix over different choices of starting layer $m$ and last layer $n$. The results are given in \autoref{fig:recoupling_rate}.

\autoref{fig:recoupling_rate}(a) is the Optimal Transport matrix $M$ for $(n, m) = (0, 6)$. A low recoupling ratio of 0.53 signals that one might encounter a lot of issues related to flow crossing. In comparison, in \autoref{fig:recoupling_rate}(b), which gives the OT matrix $M$ for $(n, m) = (6, 18)$, with a zero recoupling ratio, the existing pairings for this batch of data are already optimal.  Therefore, the recoupling ratio predicts that the quality of the latent flow layer when replacing layers 6 to 18 should be better than replacing layers 0 to 6, even though the flow over twice as many layers must be learned.  

Prior studies have shown that in a typical transformer LLM, the first and the last few layers often apply transformations to the hidden state in a way qualitatively different from those by the middle layers. In \autoref{fig:recoupling_rate}(c), the recoupling ratio over all possible choices of $m$ and $n$ is given.  We can see that the recoupling ratio from our experiments agrees with the past observation for the early layers. By exhibiting quite serious flow crossing issues, the first few layers in particular look very different from other layers.  








\vspace{-0.2cm}
\subsection{Distillation Quality on The Pile}
\vspace{-0.2cm}
In this section, we first empirically demonstrate the convergence behavior of LFT when trained via Standard Flow Matching (LFT-SFM) in Algorithm \ref{algo:standard_flow_matching} and Flow Walking (LFT-FW) in Algorithm \ref{algo:FW}. We then compare these methods against two baselines: (i) layer-skipping, which directly treats the source hidden state as the target, and (ii) a regression model \cite{gromov2025unreasonableineffectivenessdeeperlayers}, which uses a single transformer layer (initialized from the teacher’s source layer) trained with an MSE loss to map source to target hidden states.

We distill on 2.6 billion tokens from The Pile by replacing either layers 6–12 (25 \% of parameters) or layers 6–18 (50 \% of parameters) of Pythia-410m with a single flow-matching layer. As shown in \autoref{fig:training_curve}, both LFT-SFM and LFT-FW converge rapidly and substantially outperform naive layer-skipping. We then perform inference at multiple discrete time points and compare to the regression baseline in \autoref{table:comparison_selected} (see \ref{sec:inference-time-points} for analysis on time-point selection). Although LFT-SFM performance exceeds layer-skipping, it fails to match the regression model’s performance—likely because Standard Flow Matching struggles to learn fine-grained velocity fields when paired trajectories are closely clustered in latent space (\ref{sec:fw-paired-data-flow-matching}). In contrast, Flow Walking consistently outperforms both baselines in terms of latent-state matching and downstream language modeling. Notably, LFT-FW with $k=1$ achieves parity with the regression model and outperforms LFT-SFM with $k=8$, suggesting that Flow Walking’s implicit velocity estimates more accurately guide the model toward its target hidden state.


\begin{table}[t!]
\vspace{-0.5cm}
\centering
\scalebox{0.7}{%
\begin{tabular}{lcccc}

\cmidrule[\heavyrulewidth](lr){1-3}\cmidrule[\heavyrulewidth](lr){4-5}
 & \multicolumn{2}{c}{\textbf{Latent Matching}} & \multicolumn{2}{c}{\textbf{Language Modeling}} \\
\cmidrule(lr){2-3} \cmidrule(lr){4-5}
\textbf{Method} & \textbf{KL$_{x||\hat{x}}$} & \textbf{NMSE} & \textbf{KL$_{P||Q}$} & \textbf{PPL} \\
\cmidrule(lr){1-3}\cmidrule(lr){4-5}
Pythia-410m & - & - & - & 9.02 \\
\cmidrule(lr){1-3}\cmidrule(lr){4-5}
Skip one layer (layer 6) & 0.052 & 0.180 & 0.257 & 11.43\\
Skip two layer (layer 6, 7)  & 0.307 & 0.266 & 0.529 & 15.27\\
Skip three layer (layer 6, 7, 8)     & 1.157 & 0.307 & 0.932 & 22.59\\
\cmidrule(lr){1-3}\cmidrule(lr){4-5}
\textit{Replacing Layer 6-12} & & & &\\
Regression          & 0.042 & 0.010 &  0.305 & 12.45  \\
LFT - SFM ($k=1$)    & 0.076 & 0.015  & 0.409 & 14.04\\
LFT - SFM  ($k=3$)   & 0.076 & 0.015  & 0.407 & 13.98 \\

LFT - FW ($k=1$) & 0.050 & 0.012   & 0.300 & 12.42\\
LFT - FW ($k=3$) & \textbf{0.032} & \textbf{0.007}  & \textbf{0.254} & \textbf{11.86}\\

\cmidrule(lr){1-3}\cmidrule(lr){4-5}
\textit{Replacing Layer 6-18} & & & \\
Regression               & 0.156   & 0.024  & 0.864 & 21.97 \\
LFT - SFM ($k=1$)     & 0.282  & 0.040 & 1.216 & 31.35\\
LFT - SFM ($k=3$)     & 0.295  & 0.042 & 1.232 & 31.91\\

LFT - FW ($k=1$) & 0.160   & 0.031 & 0.838 & 21.82\\
LFT - FW ($k=3$) & \textbf{0.115}   & \textbf{0.021} & \textbf{0.736} & \textbf{19.92}\\

\cmidrule[\heavyrulewidth](lr){1-3}\cmidrule[\heavyrulewidth](lr){4-5}

\end{tabular}
}
\vspace{0.3cm}
\caption{\textbf{Comparison of transformer‐layer reduction methods.} We evaluate LFT at multiple discrete time points $k$; \textbf{boldface} highlights the top‐performing method within each group of layer‐replacement experiments.}
\label{table:comparison_selected}
\vspace{-0.8cm}
\end{table}


\vspace{-0.2cm}
\subsection{Effect of discrete time points on inference performance}
\label{sec:inference-time-points}
\vspace{-0.2cm}

    

The number of discrete time points $k$ is a key hyperparameter of the LFT at inference time (see Algorithm \ref{algo:LFT-inference}). While minimizing the total number of learnable parameters is crucial, it is equally important to choose a small $k$, since each additional time point incurs extra FLOPs during inference. In \autoref{fig:inference_metric_vs_step}, we report both the $\text{KL}_{x||\hat{x}}$ and NMSE of LFT as functions of $k$.  
For LFT-SFM, both KL and NMSE decrease as k is reduced, implying that early velocity estimates steer more correctly the hidden state toward its target hidden state. 
As to LFT-FW, it attains its best performance at $k=3$, matching the three‐step integration in \autoref{eq:FW_general_form}.
As $k$ approaches 1, KL divergence rises sharply, indicating that the implicit velocity estimation at $t=0$ is inaccurate and must be refined through multi‐step corrections. Interestingly, despite never training with more than three integration steps, LFT-FW extrapolates robustly to $k=8$, showing only minor degradation.

\vspace{-0.2cm}
\section{Related Work}
\vspace{-0.2cm}



\textbf{Theory of Transformer.}
Transformer architectures \cite{vaswani2017attention} are the foundation of LLMs, enabling strong performance across various NLP tasks. As model scale increases, understanding their internal structure becomes essential. Recent works explore the evolution of representations across layers, attention distributions, and layer-level modularity. For example, “Frankenstein” models recombine layers from different checkpoints to study knowledge localization and compatibility \cite{matena2022merging, ilharco2023editing, kim2023solar}. Other research investigates latent trajectory and attention alignment across layers \cite{wang2024latent}, offering insights into the compositional behavior of LLMs.

\textbf{Efficient Transformer.}
Improving transformer efficiency has been approached through pruning, parameter sharing, and distillation. Magnitude-based pruning \cite{gordon2020compressing} and structured approaches \cite{zafrir2021prune} have shown that substantial parameter reduction is possible with limited performance drop \cite{lu2024reassessing}. Parameter-sharing architectures like ALBERT \cite{lan_albert_2020} compress models by reusing weights across layers. Distillation methods such as DistilBERT \cite{sanh2019distilbert} and TinyBERT \cite{jiao2019tinybert} transfer knowledge from larger models to smaller ones through intermediate supervision and multi-stage training.

\textbf{Flow matching for image generation.}
Flow-based models have emerged as efficient alternatives to diffusion in generative modeling. Flow Matching (FM) \cite{lipman2022flow} learns continuous velocity fields for optimal transport and has been extended with Rectified Flow \cite{liu2022rectified} and Conditional Flow Matching \cite{li2023diffusion} to improve training stability and sample quality. Recent works \cite{frans_one_2024} demonstrate high-quality one-step generation by learning direct endpoint mappings, highlighting FM’s potential in reducing generation steps.

\textbf{Diffusion and flow matching models for language modeling.}
Recent efforts have adapted diffusion and FM frameworks for language generation. Diffusion-based models such as GENIE \cite{lin2023text} apply denoising processes to continuous token embeddings, while AR-Diffusion \cite{wu2023ar} introduces autoregressive properties for better sequence modeling. FM-based approaches include FlowSeq \cite{hu2024flowseq}, which improves sampling efficiency, and CaLMFlow \cite{he2024calmflowvolterraflowmatching}, which applies Volterra integral formulations to better align continuous and discrete modeling in language.

\vspace{-0.2cm}
\section{Discussions and open problems}
\vspace{-0.2cm}
\label{sec:discussion}
Speculative decoding refers to the use of a small draft model to accelerate the inference of a full-sized model. As the size of a straight flow transformer can be order-of-magnitude smaller than its teacher model, in our opinion it is natural to use the straight flow transformer as the draft model in a speculative decoding set up. Given the KL-distance result that was presented in our work, the performance of such a configuration is quite promising.  

A transformer has a fixed computational cost with a fixed generation quality. In contrast, a straight flow transformer allows for dynamically varying the number of steps, i.e. the computational investment, on a sentence-by-sentence, or even a token-by-token basis. Thus one might contemplate whether and how to reach an optimal trade-off between the number of steps used and the corresponding quality. However, how to self- or cross-estimate the quality of the output logits of a given token at the output of a transformer is still an open problem.

Lately, state-space models and recurrent models have emerged as candidates to substitute for transformers due to their constant computational complexity of handling context. We suspect that if the method in our work is applied, one can create straight flow RWKV, straight flow MAMBA, straight flow xLSTM, etc. Unfortunately, at the moment we are not properly setup to train SOTA state-space or recurrent models. 

\vspace{-0.2cm}
\section{Future Work}
\vspace{-0.2cm}

\textbf{Flow Untangle.} In our current approach, we have retained the input and output layers of the source transformer unchanged. It is well established that arranging flow matching pairs to minimize flow crossing significantly enhances performance. We propose optimizing the input and output layers to minimize flow crossing, conditioned on the input embedding and output logit. Although we are actively working on this optimization, results are not yet available at the time of writing. Successfully implementing this strategy could reduce the number of layers by an order of magnitude or more, thereby improving inference efficiency.

\textbf{Architectural Search and Initialization Method.} In this paper, we have distilled the interior layers of the source transformer into straight flow layers, presenting a principled pruning method. Post-pruning, we aim to investigate the training dynamics if pretraining is continued. This approach has the potential to outperform its parameter class, offering significant benefits for training by reducing the number of layers by an order of magnitude or more.

\textbf{Scale Up to Larger Models} Due to limited computational resources, our experiments have been constrained to the Pythia model. In our future work, we plan to scale up our experiments, but achieving this requires collaboration with researchers who have access to larger computational power. Extending our approach to state-of-the-art models, such as the larger OLMO, is a crucial next step in validating and enhancing our findings.

\textbf{Training Flow-Replaced Transformers from Scratch.} An open question is whether the resulting shallow transformer, once flow-replaced and structurally simplified, can be trained from scratch without first pretraining a full-depth model. Prior observations in diffusion literature suggest that one-step generators trained from scratch rarely match the performance of distilled ones. A similar pattern may hold in language modeling. This motivates further investigation into curriculum training, staged flow replacement, or leveraging lightweight teacher guidance during early training stages.

\medskip
\bibliographystyle{plain}
\bibliography{ref}



\appendix

\ifarxiv
\fi

\end{document}